\documentclass[10pt]{article}

\PassOptionsToPackage{numbers, compress, sort}{natbib}

\usepackage{hyperref}
\usepackage{url}

\usepackage{amsmath}
\usepackage{amssymb}
\usepackage{amsthm}
\usepackage{bm}
\usepackage{booktabs}       
\usepackage{subfigure}
\usepackage{caption}
\usepackage{wrapfig}
\usepackage{colortbl}

\usepackage{thmtools, thm-restate}
\usepackage{geometry}

\usepackage{verbatim}
\usepackage{bbm}
\usepackage{systeme}
\usepackage{times}  
\usepackage{helvet}  
\usepackage{courier}  
\usepackage{url}  
\usepackage{graphicx}  
\usepackage{natbib}
\usepackage[utf8]{inputenc} 
\usepackage[T1]{fontenc}    
\usepackage{lmodern}
\usepackage{hyperref}       
\usepackage{url}            
\usepackage{booktabs}       
\usepackage{amsfonts}       
\usepackage{amsmath}  
\usepackage{amsthm}
\usepackage{amssymb}  
\usepackage{nicefrac}       
\usepackage{microtype}      
\usepackage{bbm}
\usepackage{tikz}
\usetikzlibrary{arrows}
\usetikzlibrary{positioning}
\tikzset{
  treenode/.style = {align=center, inner sep=0pt, text centered,
    font=\sffamily},
  arn_n/.style = {treenode, circle, black, font=\sffamily\bfseries, draw=black,
    fill=white, text width=1.5em},
  arn_r/.style = {treenode, circle, black, font=\sffamily\bfseries, draw=black,
    fill=white, text width=1.0em},
  arn_x/.style = {treenode, rectangle, draw=black,
    minimum width=0.5em, minimum height=0.5em}
}
\usepackage{thmtools,thm-restate}
\usepackage{parskip}
\usepackage{authblk}
\usepackage{subcaption}
\usepackage[linesnumbered,ruled,noend]{algorithm2e}
\usepackage{algorithmic}
\usepackage[subtle]{savetrees}
\usepackage{mathtools}
\usepackage{dsfont}
\usepackage[many]{tcolorbox}
\usepackage{hyperref}
\usepackage{wrapfig}
\usepackage{graphicx}

\usepackage{basic}

\usepackage{etoolbox}

\makeatletter
\newcommand{\changeoperator}[1]{%
  \csletcs{#1@saved}{#1@}%
  \csdef{#1@}{\changed@operator{#1}}%
}
\newcommand{\changed@operator}[1]{%
  \mathop{%
    \mathchoice{\textstyle\csuse{#1@saved}}
               {\csuse{#1@saved}}
               {\csuse{#1@saved}}
               {\csuse{#1@saved}}%
  }%
}
\makeatother

\let\oldnl\nl
\newcommand{\nonl}{\renewcommand{\nl}{\let\nl\oldnl}}

\newif\ifsinglecolumn
\singlecolumntrue

\title{Evaluating Language Model Context Windows:\\ A ``Working Memory'' Test and Inference-time Correction}

 \author[$\dagger$]{Amanda Dsouza}
 \author[$\dagger$]{Christopher Glaze}
 \author[$\diamond$]{Changho Shin}
 \author[$\dagger \diamond$]{Frederic~Sala}

 \affil[$\dagger$]{Snorkel AI}
 \affil[ ]{\footnotesize{\texttt{\{adsouza, chris.glaze\}@snorkel.ai}}}
 \affil[$\diamond$]{University of Wisconsin-Madison}
 \affil[ ]{\footnotesize{\texttt{\{cshin23, fredsala\}@wisc.edu}}}
\date{}

\begin{document}
    \maketitle

\begin{abstract}
Large language models are prominently used in real-world applications, often tasked with reasoning over large volumes of documents. 
An exciting development in this space is models boasting extended context capabilities, with some accommodating over 2 million tokens. 
Such long context model capabilities remain uncertain in production systems, motivating the need to benchmark their performance on real world use cases. 
%
We address this challenge by proposing SWiM, an evaluation framework that addresses the limitations of standard tests. 
Testing the framework on eight long context models, we find that even strong models such as GPT-4 and Claude 3 Opus degrade in performance when information is present in the middle of the context window (lost-in-the-middle effect). 
Next, in addition to our benchmark, we propose \emph{medoid voting}, a simple, but effective training-free approach that helps alleviate this effect, by generating responses a few times, each time randomly permuting documents in the context, and selecting the medoid answer.
We evaluate medoid voting on single document QA tasks, achieving up to a 24\% lift in accuracy.
Our code is available at \href{https://github.com/snorkel-ai/long-context-eval}{https://github.com/snorkel-ai/long-context-eval}.
\end{abstract}
\section{Introduction}

\begin{wrapfigure}{R}{0.45\textwidth}
	\centering
\includegraphics[width=0.45\textwidth]{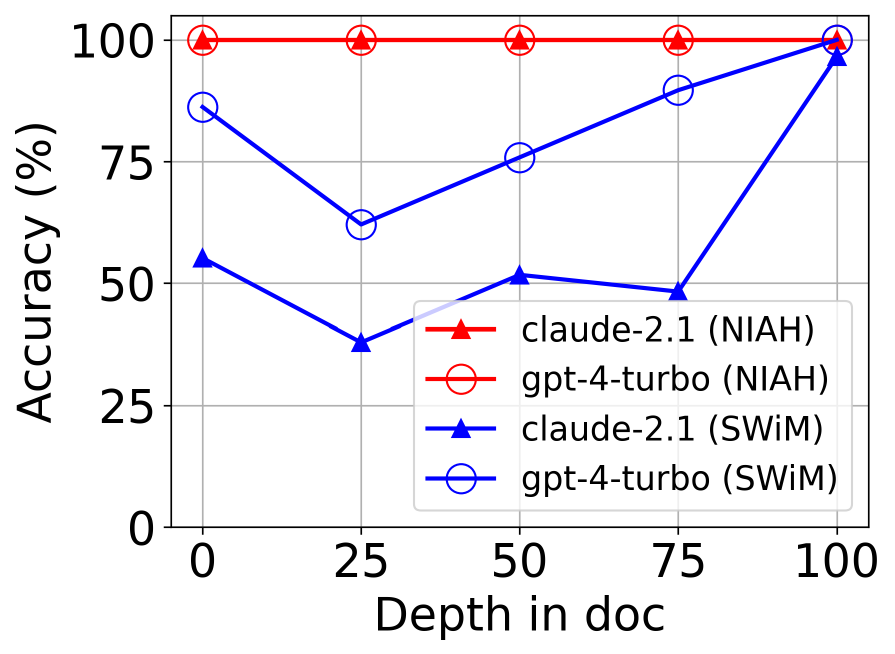}
\caption{\small  Results of NIAH (answering a synthetic “What is the best thing to do in San Francisco?” needle on Paul Graham essays as the haystack), alongside SWiM on a QA task. GPT-4 and Claude 2.1 obtain perfect scores on the NIAH test, at all document depths. But a more realistic QA task on narrative content using SWiM reveals the typical “lost-in-the-middle” effect.}
    \label{fig:niah_swim_comparison}
\end{wrapfigure}

Real-world applications increasingly use large language models (LLMs) to analyze extensive document collections. Although new models can ingest extended contexts, some accommodating up to 2 million tokens, their actual capabilities --- whether they can effectively utilize long context input in practical applications --- remain unclear. Evaluating their performance in real-world scenarios is important to enabling these models to operate in production settings.

Evaluating long context capabilities of LLMs has so far been restricted to the popular ``needle in a haystack'' (NIAH) test \cite{LLMTest_NeedleInAHaystack}, its variants \cite{hsieh2024ruler}, or through academic benchmarks \cite{bai2023longbench,zhang2024infty}. While these may serve as useful starting points for evaluating new model releases, their relevance and applicability to real-world problems are unclear for several reasons: 
\begin{itemize}[topsep=0pt,]
    \item \textbf{Unrelated Tasks:} The NIAH test often uses unrelated \emph{toy} needles, which do not capture the intricacies of information retrieval in documents, where context and relevance are crucial.
    \item \textbf{Limited Applicability:} Academic benchmarks are often based on datasets that do not represent the nuances and complexities of documents found in real-world scenarios, such as financial reports, legal documents, or customer feedback.
\end{itemize}

\begin{itemize}
    \item  \textbf{Varying model performance:} Model performance can vary significantly depending on the task and data. For instance, experiments by \cite{hsieh2024ruler} showed that even on standard tasks/datasets, GPT-4 accuracy went from 95-100\% on NIAH / retrieval tasks to 79.7\% on word extraction tasks to 59.0\% on QA tasks.
\end{itemize}

\begin{figure*}[!t]
    \centering
    \includegraphics[width=\textwidth]{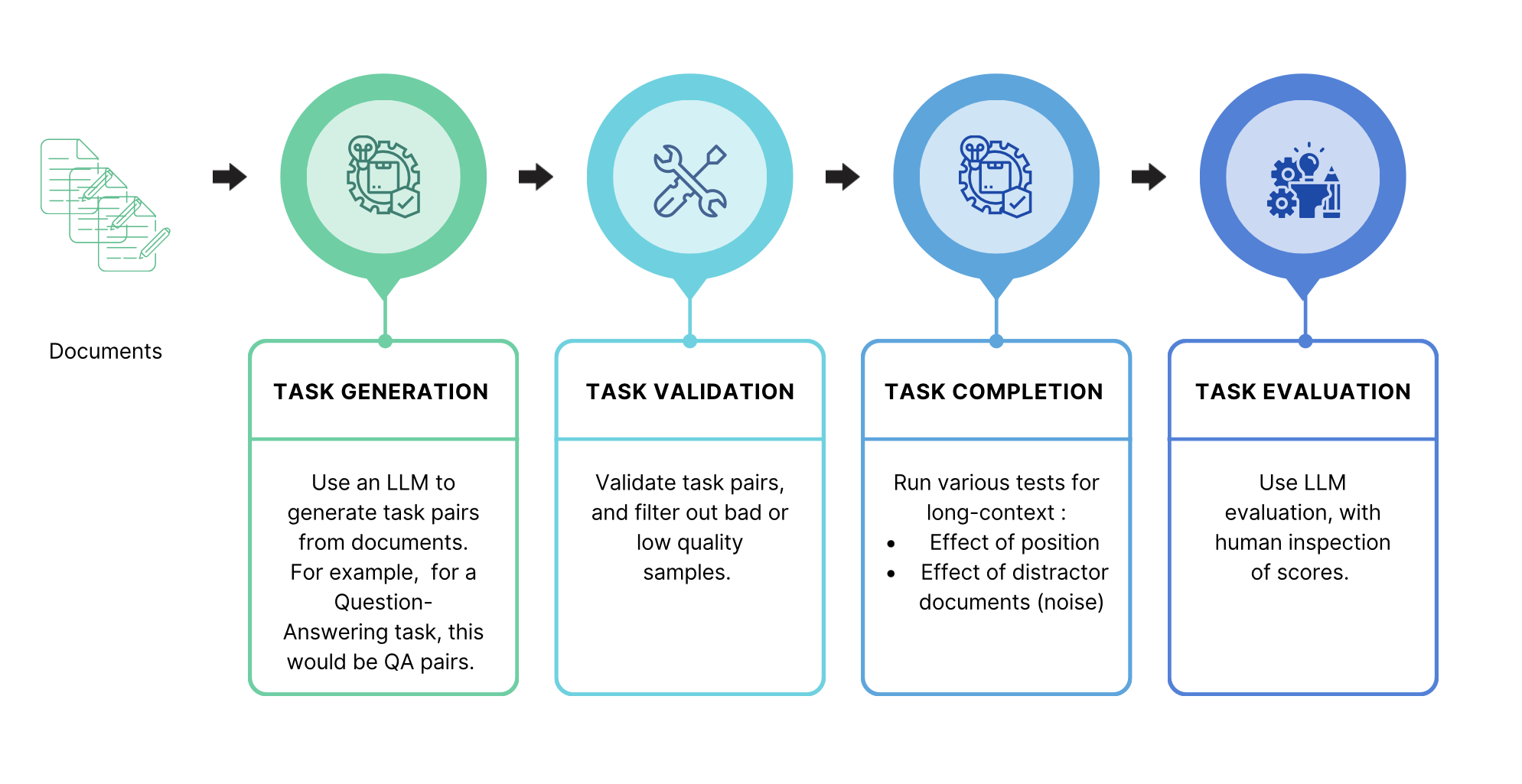}
    \caption{SWiM Framework}
    \label{fig:swim_framework}
\end{figure*}
\section{SWiM: Long Context Evaluation Framework for Real-world Tasks}

To address these problems, we propose Snorkel Working Memory Test (SWiM), an  evaluation framework which measures long context capabilities of any language model on use-case specific documents and task pairs. The test and its name are inspired by a wealth of research on human cognition suggesting that short-term memory buffering (as is required in simple recall tasks such as the NIAH) is just one component of a more complex working memory network engaged in using buffered information for more executive functions such as reasoning and goal-directed behavior 
\cite{baddeley1974workingmemory,engles2002workingmemory,cowan2008workingmemory,desposito2015workingmemory}.

The SWiM test is executed with a four-step process (Figure \ref{fig:swim_framework}) of task generation, task validation, task completion and task evaluation. Directly testing on the relevant data and tasks allows a more realistic assessment of a model’s long context capabilities for unique applications. Our code is available at \href{https://github.com/snorkel-ai/long-context-eval}{https://github.com/snorkel-ai/long-context-eval}.

We use SWiM to test eight long context models (including from Open AI, Anthropic, Google, and Mistral), for their effective context length, and the effect of position. Most long context models are ineffective at retrieving information in the middle of the context window (confirming "lost-in-the-middle" effect \cite{liu2024lost}). To mitigate this effect, we additionally propose a simple algorithmic approach, \emph{medoid voting}, a straightforward yet effective training-free method that improves performance.

The framework takes the following four steps.

\begin{enumerate}
    \item \textbf{Task generation}: In order to benchmark long context models on use-case specific data, we automate the creation of the task using a large language model. We create a Question-Answering task that automatically creates QA pairs on user documents using a large language model.
    \begin{tcolorbox}[top=3pt,bottom=3pt,left=3pt,right=3pt]

    \textbf{Task generation prompt}
        
        Ask a factoid question given only the context provided, that can be answered in a few words. Answer the question given the context. Format your result in a JSON object with keys `question' and `answer'.
        
         Context: \{context\}
         
         Result:
    \end{tcolorbox}
    
    \item \textbf{Task validation}:  Using language models to replace human intensive tasks such as dataset creation and evaluation has become commonplace as strong models continue to be released. They are however, prone to errors and without a validation step allows these inaccuracies to trickle down the pipeline, leading to compounding errors and ultimately influencing reported results.
    
     While prompting and adding guardrails help to some extent, we find that the validation step cannot be circumvented, and a human-in-the-loop approach is best at ensuring good data at the input and output.

    \item \textbf{Task Completion}: SWiM benchmark supports document QA retrieval, and tests specifically for (a) the effect of position on retrieval accuracy, (b) the effect of context size (the level of noise) on retrieval performance.
    \begin{enumerate}
        \item \textbf{Effect of position}: \cite{liu2024lost} showed that models are better at retrieving information from the top or bottom of a long context, while performance greatly reduces when it is contained in the middle. They refer to this phenomenon as the “lost-in-the-middle” effect. In a similar vein, the popular “needle-in-a-haystack” test measures how position affects model performance by injecting a synthetic needle into a set of essays.
        
        We conduct a similar test to measure how positioning the document within the context window affects retrieving the answer from it. This is similar to the “needle-in-a-haystack” test except over user documents and realistic needles, rather than the synthetic test. This also effectively tests the “lost-in-the-middle” effect, although the experiment setup is slightly different. Liu et al. test on a constant number of 10, 20 and 30 distractor documents (documents that do not contain the answer, but are in the same domain as the answer document), that may not reach the full context window of the model, as well as present the distractor documents in order of decreasing relevance. In contrast,  SWiM tests a model on its full context window, with distractor documents shuffled at random.
        To measure the effect of position on retrieval performance, we follow the procedure in \ref{alg:testing_document_position}. In our experiments, we test varying depths of the true response at 0, 25, 50, 75, and 100\% depths.

        \item \textbf{Effect of context size:} Starting with just the answer document in context (0\% noise), distractors are successively added, increasing the capacity to 25, 50, 70 and 100\% of the context window and model responses generated to test the effective context length of the model in the presence of distractors. To measure the effect of context window used on retrieval performance, we follow the procedure in \ref{alg:testing_long_context_and_rag}.
    \end{enumerate}
    
    \item \textbf{Evaluating responses}:  We use LLM-as-a-judge to evaluate the responses. On the Single Document QA task, we use the following prompt.
    \begin{tcolorbox}[top=3pt,bottom=3pt,left=3pt,right=3pt]

    \textbf{Evaluation prompt}
    
       For the question provided, return a JSON object with a `correct' key as a boolean, if the given answer and gold standard answer are the same in meaning (may be worded differently).
               
        Question: \{question\}
        
        Answer: \{answer\}
        
        Gold standard answer: \{gold\_answer\}
        
        JSON:
    \end{tcolorbox}
    
As with the task generation step, evaluation with a language model is error prone. While the task is a relatively simple one (check if both answers are correct in semantics), there are cases where we find the LLM judge predicts the wrong score. In some cases, the responses contain additional detail, which requires validating the additional piece of information from the document. Furthermore, we find that LLM judge responses can be inconsistent, even with greedy decoding, giving different scores on the same (or very similar) responses.

\end{enumerate}

\begin{algorithm}[t!]
	\caption{\bf SWiM test of document position}
	\begin{algorithmic}[1]\label{alg:testing_document_position}
		\STATE 
            Get $m$ documents $\mathcal{D}=\{D_1, D_2, \ldots D_m\}$.\footnote{In practice, the document set might not fit the context window. In that case, we randomly choose the subset of the document set that includes the answer documents}
        \STATE Get (question, answer, answer document index) triplets $(Q_1, A_1, i_1), \ldots (Q_n, A_n, i_n)$.
            \FOR{$k \gets 1$ to $n$}
                \FOR{ each position $p$ in (0, 25, 50, 75, 100)}
                \STATE Get $m-1$ distractor documents $\mathcal{D}_{-k}=\mathcal{D}-D_{i_k}$
                \STATE Create context $\mathcal{C}$ with Shuffled $\mathcal{D}_{-k}$ and answer doc $D_{i_k}$ fixed at position $p\%$.
                
                \STATE Generate response to $Q_k$ given context $\mathcal{C}$
                \ENDFOR
            \ENDFOR
	\end{algorithmic}
\end{algorithm}

\begin{algorithm}[t!]
	\caption{\bf SWiM test of context size}
	\begin{algorithmic}[1]\label{alg:testing_long_context_and_rag}
		\STATE Get $m$ documents $\mathcal{D}=\{D_1, D_2, \ldots D_m\}$
            \STATE Get (question, answer, answer document index) triplets $(Q_1, A_1, i_1), \ldots (Q_n, A_n, i_n)$.
            
            \FOR{$k \gets 1$ to $n$}
            \STATE $D^* = D_{i_k}$   
            \FOR{the distractor percentage $p\%$ in (0, 25, 50, 75, 100)}         \STATE Sample $p\%$ documents of $\mathcal{D}$ to construct distractor document set $\tilde{\mathcal{D}}$
                \STATE Construct context $\mathcal{C}=$Shuffle$(\{D^*\}\cup \tilde{\mathcal{D}})$
                \STATE Generate response to $Q_k$ given context $\mathcal{C}$
                \ENDFOR
            \ENDFOR
	\end{algorithmic}
\end{algorithm}






\section{Medoid Voting: a simple solution to lost-in-the-middle effect}
We propose an efficient solution to the lost-in-the-middle effect. Since we know models are generally less effective at retrieving information from some positions than others, it would help if we positioned the answer documents so that they are in the right places. Of course, we do not know what the right place is a priori. A possible solution to this problem is to run the task completion a few times, each time randomly permuting documents in the context, and then use a selection criteria to pick the best response. We consider the medoid response (response with the least dissimilarity to all other responses) in the embedding space as our selection criteria. The detailed procedure is described in Algorithm \ref{alg:medoid_voting}.

\begin{algorithm}[t!]
	\caption{\bf Medoid voting: SWiM's correction of lost-in-the-middle effect}
	\begin{algorithmic}[1]\label{alg:medoid_voting}
		\STATE \textbf{Input:}
		  Document set $D$, query $q$, \# of permutations $k$, language model $f$, embedding model $g$
            \STATE Sample random permutations $\sigma_1, \ldots, \sigma_k$ for $D$
            \STATE Get LLM ouputs $\hat{y}_i=f([\sigma_{i}(D), q]), i=1,\ldots,k$
            \STATE Get embeddings of LLM ouputs $e_i=g(\hat{y}_i), i=1,\ldots,k$
            \STATE Get medoid in embedding space $\hat{y} = \hat{y}_{i^*}$, where $i^*= \arg\max_{i \in \{1, \ldots, k\}} \sum_{j=1}^k \cos(e_i, e_j)$.
            \STATE \textbf{Return } $\hat{y}$
	\end{algorithmic}
\end{algorithm}

\begin{figure*}[t!]
	\centering
\includegraphics[width=0.85\textwidth]{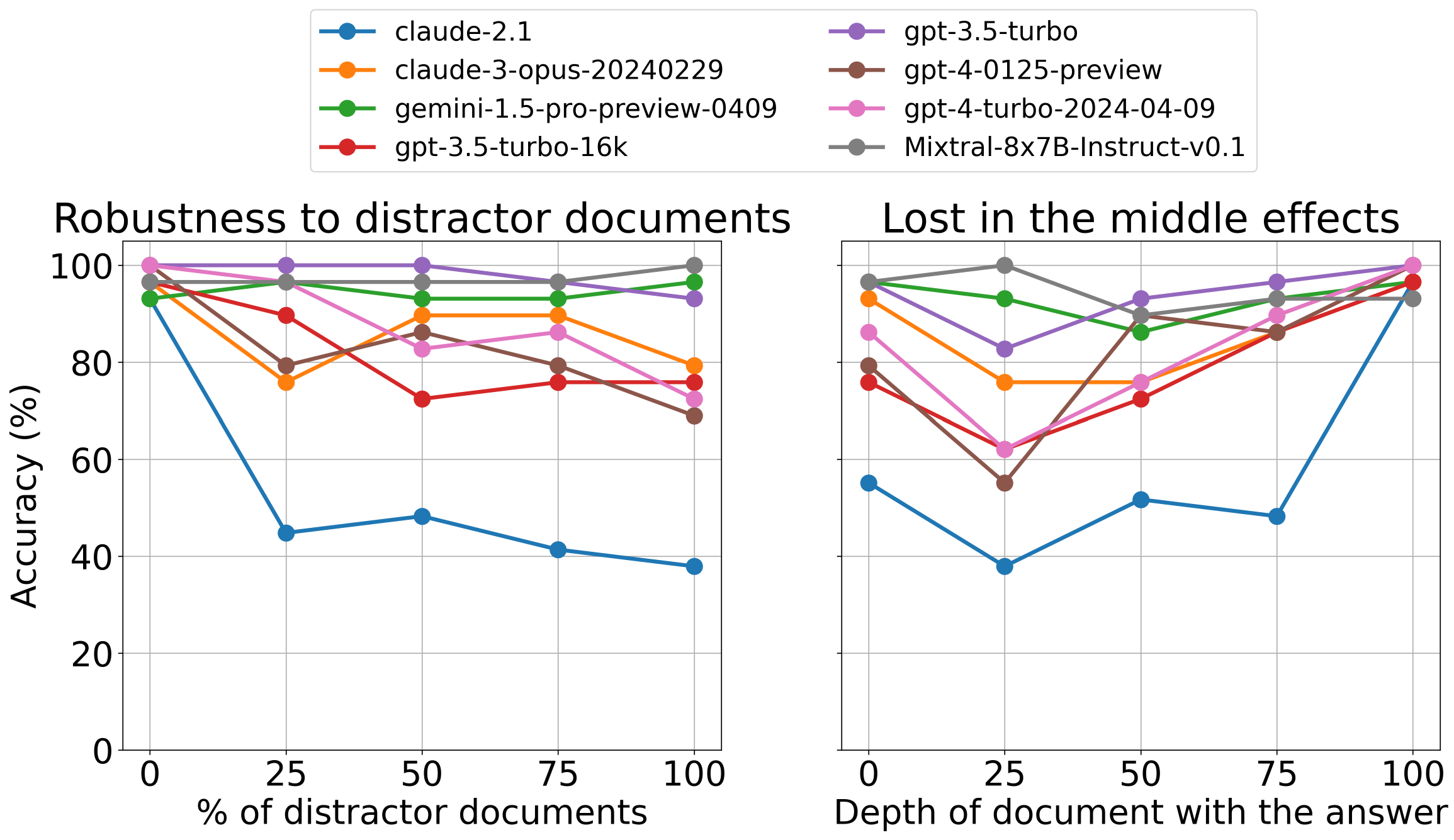}
\caption{\small More analyses on LLMs with SWiM framework. (Left) shows that not all models utilize their long context windows effectively, though their context window lengths are enough to include entire documents. (Right) reveals "Lost-in-the-middle" effect is significant and common across many LLMs.}
    \label{fig:main_exp}
\end{figure*}

\section{Experimental Results}
For our experiments, we create QA pairs using GPT-4, from the synthetically generated Huggingface Cosmopedia story\_forum dataset \cite{benallal2024cosmopedia}, treating each instance as a document. We use SWiM to test long context models by Open AI, Anthropic, Google, and Mistral on the Document QA task. Specifically, we evaluate LLMs' robustness to distractor documents and ``lost-in-the-middle'' effects. Finally, we validate the effectiveness of medoid voting. 

\subsection{Evaluation on robustness to distractor documents}
We evaluate robustness to distractor documents given in the context. This investigates how well LLMs can extract the necessary information without being distracted to irrelevant context.

\noindent \textbf{Setup.}
We test single document QA performance with an increasing number of documents that do not contain the answer (distractor documents). Starting with only the answer document, distractors are added to fill up the context window to 25, 50, 75 and 100\% of its capacity\footnote{For simplicity we use tiktoken to count tokens. For non OpenAI models, this means that we use slightly lower context sizes than the reported size.
}, and the set is shuffled within the context window for response generation.

\noindent \textbf{Results.} Figure \ref{fig:main_exp} (left) shows the experiment results. Unsurprisingly, as the number of distractors increase, performance degrades. But this degradation is not uniform across models. Among models with a long context window (1M tokens), Gemini-1.5-Pro does extremely well to handle noise. It is also unsurprising that smaller context length models (such as GPT-3.5 Turbo and Mistral-8x7B-Instruct) are more effective at using their context lengths compared to larger context length ones. Analyzing incorrect responses, we find that many of these are due to how documents are positioned in the context, which we discuss next.


\begin{figure*}[t!]
\centering
  \includegraphics[width=\linewidth]{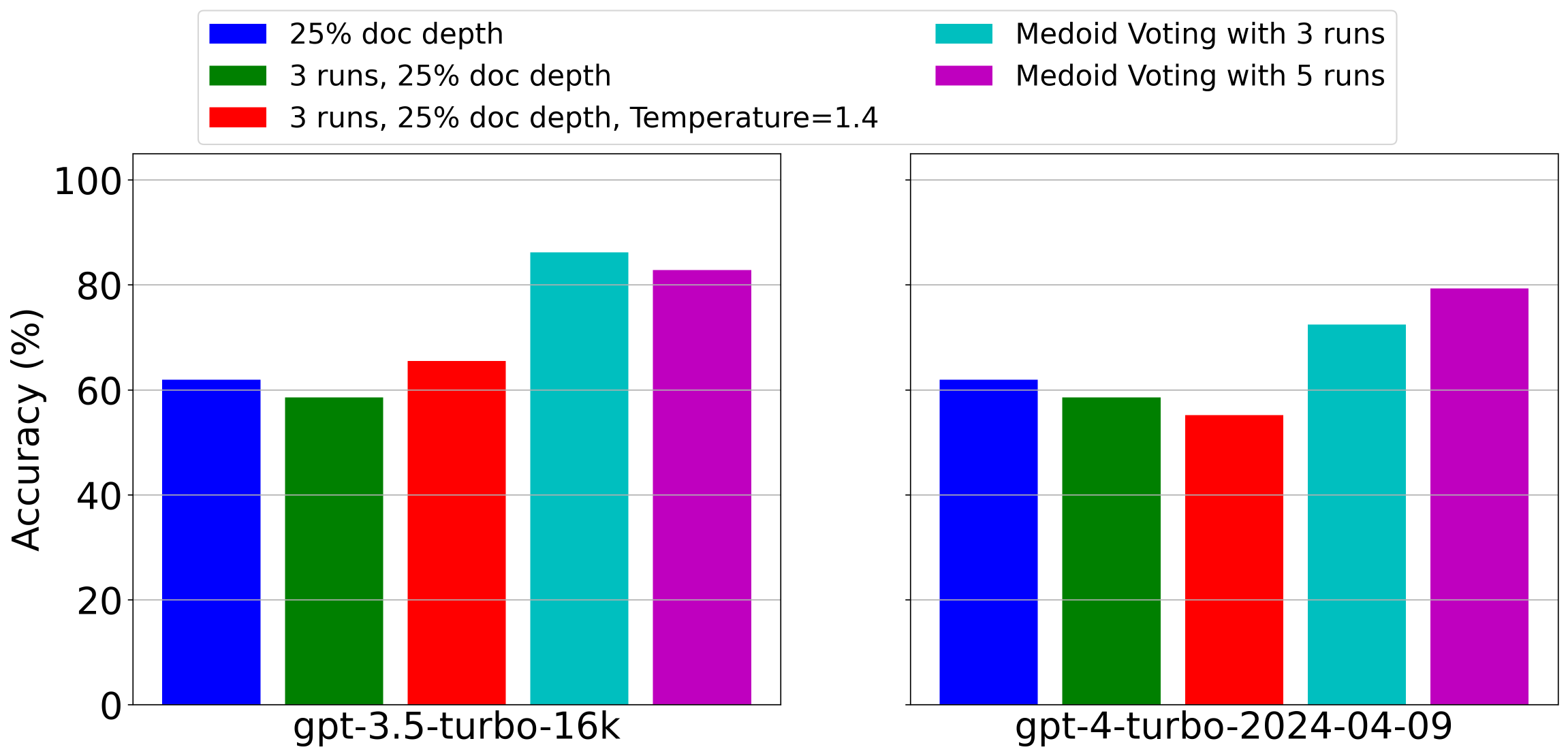}
  \label{fig:medoid_voting_left}
\caption{\small Medoid voting can easily smooth out the "lost-in-the-middle" effect.}
\label{fig:medoid_voting}
\end{figure*}


\subsection{Evaluation on robustness to document position}
We evaluate long-context LLMs' robustness to document position. It is well known that position affects the retrieval performance of long context LLMs, but the effect varies depending on benchmarks and models \cite{liu2024lost,zhang2024infty}. We expect that lost-in-the-middle effect is common to long context models.

\noindent \textbf{Setup.}
We use SWiM to test eight models on the single document QA task. We place the answer document at the 25, 50, 75 and 100\% positions in the context length and fill up the rest of the context window with randomly shuffled distractor documents. Note that since we first compute the number of tokens of documents to fit the context window of a model, the number of documents used for a model may differ based on the model's context window size. Across the position tests for a given model, the set of documents is kept fixed, to reduce confounding effects of distractors on the performance.

\noindent \textbf{Results.}
Figure \ref{fig:main_exp} (left) shows the experiment results. Most models exhibited a degradation in performance at the 25\% $\sim$ 75\% depth and showed their best performance when the answer is located at 0\% depth or 100\% depth. The observed performance degradation of long context models in the middle poses significant implications for real world applications. 

\begin{wrapfigure}{R}{0.45\textwidth}
	\centering
\includegraphics[width=\linewidth]{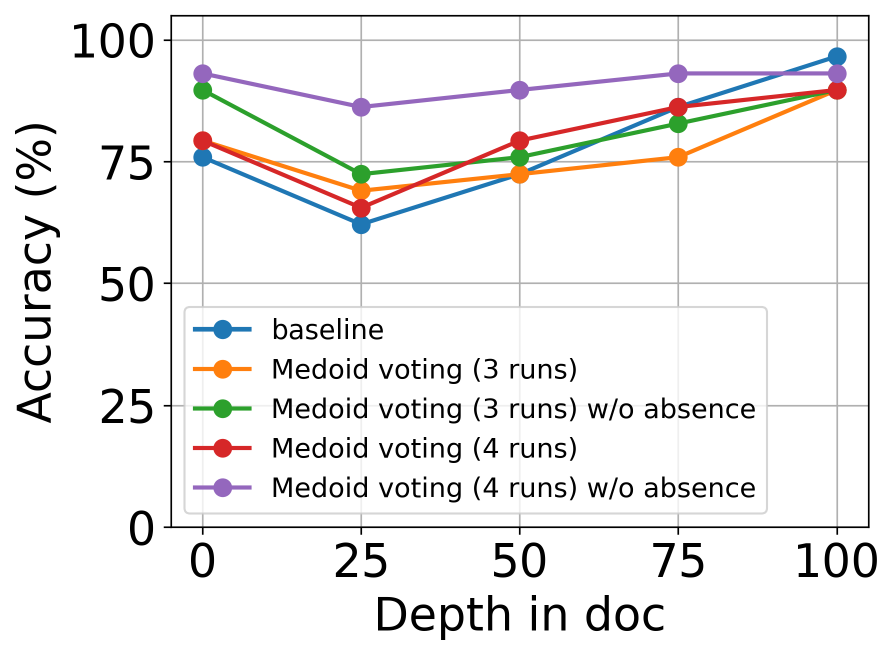}
  \caption{\small Voting accuracy depending on the depth of document with answer}
  \label{fig:medoid_voting_no_absence}
  \vspace{-12mm}
\end{wrapfigure}

Additionally, this result implies that the NIAH test, while a good way to quickly test new models, is often not a good indicator of real world performance. This is evident in Claude-2.1’s results, where NIAH results showed strong performance when the needle was at the top and bottom of the context, but we do not find the same behavior when the document is at the top of a long context.

\subsection{Medoid voting}

\noindent \textbf{Setup.} We tested medoid voting on two models that observed the lost-in-the-middle effect, GPT-4-Turbo and GPT-3.5-Turbo-16k. To isolate the impact of the medoid voting, we conducted a control experiment that generates multiple runs keeping document position constant at the unfavorable 25\% document depth, with model stochasticity as the only source of variation. We used  temperatures of 0.7 (default) and 1.4 (a high temperature to induce higher variance in responses). This lets us test the specific effect that varying position has over other forms of variation in generating the final response.


\noindent \textbf{Results.}
Figure \ref{fig:medoid_voting} shows the results. We found medoid voting to be effective, even with as few as 3 runs. Results show the medoid voting method surpasses performance over both the baseline and controls, on both models; a 17.3 point lift on GPT-4-Turbo, and 24.2 point lift on GPT-3.5-Turbo-16k. This underscores medoid voting's superiority—a straightforward, training-free strategy that can be used to enhance performance. 

We additionally found that medoid voting might compromise the best case performance by averaging effect. However, we easily address this issue by filtering absence outputs ("No information found in context"), assuming the proper information can be found in the context. Figure \ref{fig:medoid_voting_no_absence} shows the analysis and further improvement by absence filtering.

\section{Related Work}
\noindent \textbf{Existing long context benchmarks.}  As the effectiveness of in-context learning \cite{brown2020incontext} has been studied, benchmarks to evaluate long-context capacity of LLMs have been actively developed.
Long Range Arena \cite{tay2020long} evaluates LLMs with the context length from 1K to 16K tokens, including various data types and modalities such as text, images, and mathematical expressions.
The needle in a haystack \cite{LLMTest_NeedleInAHaystack} (NIAH) test evaluates LLMs' capability to retrieve answers from the given context (essays). RULER benchmark \cite{hsieh2024ruler} expands upon the NIAH test to include more variations such as multi-hop tracing and aggregation. LongBench \cite{bai2023longbench} provides a benchmark with single-doc QA, multi-doc QA, summarization,
few-shot learning, code completion, and synthetic tasks in English and Chinese. $\infty$Bench \cite{zhang2024infty} provides datasets and pipelines to evaluate LLMs with 100K+ context in English and Chinese.
While these have been useful for evaluating long-context capability of LLMs, they lack customizability---their tasks might not be directly relevant to the problems encountered in business applications. Our framework, instead, provides an end-to-end customizable evaluation pipeline.
\section{Conclusion}
We propose the SWiM framework, which enables users to create a personalized benchmark to evaluate long context models on their data with their tasks. Experimental results suggest that SWiM can identify patterns of errors that NIAH cannot, so we strongly recommend usage of SWiM prior to any development work for specific applications. However, there is more experimental work required to draw general conclusions about the language models used in research reported here. While we saw that some models such as Gemini-1.5-Pro do extremely well handling long contexts in the setting of single document QA, they may be less effective in more complex scenarios. These complex scenarios include relevant documents that have been combined with many other similar documents; tasks requiring reasoning over multiple documents; and tasks requiring  citations to specific sources within documents. Future development of the SWiM framework should address these complex scenarios with more fine grained evaluations (e.g. including hallucination detection), and on a wider array of tasks.

\subsubsection*{Acknowledgments}
We thank Paroma Varma and Armin Parchami for their helpful feedback and discussion.

\bibliographystyle{plainnat}
\bibliography{ref}

\end{document}